# Optimizing Search Advertising Strategies: Integrating Reinforcement Learning with Generalized Second-Price Auctions for Enhanced Ad Ranking and Bidding

Chang Zhou[1]
Columbia University
New York, USA
mmchang042929@gmail.com

Yang Zhao[2,*]
Columbia University
New York, USA
* Corresponding author: yangzhaozyang@gmail.com

Jin Cao[2]
Independent Researcher
Dallas, USA
caojinscholar@gmail.com

Yi Shen[3]
University of Michigan
Ann Arbor, USA
shenrsc@umich.edu

Xiaoling Cui[4]
Independent Researcher
San Jose, USA
cathylingx1993@gmail.com

Chiyu Cheng[5]
University of California, Irvine
Seattle, USA
cypersonal6@gmail.com

*Abstract*—This paper explores the integration of strategic optimization methods in the context of search advertising, focusing on ad ranking and bidding mechanisms within e-commerce platforms. Employing a combination of reinforcement learning and evolutionary strategies, we propose a dynamic model that adjusts to varying user interactions and optimizes the balance between advertiser cost, user relevance, and platform revenue. Our results suggest significant improvements in ad placement accuracy and cost-efficiency, demonstrating the model's applicability in real-world scenarios.

*Keywords-reinforcement learning; search advertising; ad ranking; bidding strategies; optimization*

## I. Background

Search advertising is crucial in e-commerce, significantly boosting group revenue and supporting ecological balance and merchant growth. With big data and algorithm advancements shaping business strategies, technological innovation has become essential in search marketing. The ad ranking process involves advertisers bidding on keywords, and the search engine assessing ad quality and bid amounts to determine rankings, prioritizing higher-quality, higher-bid ads. This cost-per-click model benefits all stakeholders: advertisers enhance visibility and sales potential, users receive efficient, personalized recommendations, and platforms strive to maximize profits while ensuring satisfaction. Optimizing the ad sorting algorithm and pricing mechanism is vital, utilizing a policy optimization algorithm based on reinforcement learning[1] to refine the formula in various search contexts.

## II. Problem Statement

Optimizing ad ranking and bidding impacts revenue for advertisers, users, and platforms. Advertisers rely on display opportunities for promotion, users benefit from high-quality ads, and platforms drive revenue through engagement and transactions. The sorting formula balances these interests, aligning ad relevance with search results to optimize user engagement. The learning process for the sorting formula, defined by $a = A(s)$ where a denotes parameters and s represents the search context, spans beyond localized interactions to encompass global user sequences for maximal cumulative reward. Our focus on maximizing RPM (revenue per thousand impressions) considers CTR, CVR, and GMV to ensure sustainable platform profitability.

## III. Theoretical Framework

In the future, the sorting formula will have the ability to control the revenue of advertisers, users and platforms. We design the sorting formula as follows.

$$\phi(s, a, ad) = f_{a_1}(CTR) \cdot bid + a_2 \cdot f_{a_3}(CTR, CVR) + a_4 \cdot f_{a_5}(CVR, price) \quad (1)$$

Note that the equation is centered using a center tab stop

Where $a = a_i$ ($i = 1, \ldots, 5$) represents the parameters of the sorting formula, bid represents the user's bid for the advertisement ad, price represents the price of the product corresponding to the advertisement, CTR, CVR are The system predicts click probability and conversion probability. $f_{a_1}$ in the sorting formula can be considered as the expected value of the platform's income; $f_{a_2}$ considers the user's click probability and conversion probability, and is mainly used to describe the user's satisfaction; $f_{a_3}$ considers factors related to purchase and represents the Possible benefits to the plaintiff. In addition, $a_2$ and $a_4$ are used to adjust the balance relationship of the latter two factors. We use ad, ad' to represent advertisements ranked between two adjacent positions. According to GSP's deduction calculation method, the current click deduction can be calculated as

$$\text{click\_price} = \frac{\phi(s,a,ad') - (a_2 \cdot f_{a_3}(CTR,CVR) + a_4 \cdot f_{a_5}(CVR,price))}{f_{a_1}(CTR)} \quad (2)$$

## IV. METHODOLOGY

The emergence of artificial intelligence has profoundly impacted various fields, such as machine learning[2-6], natural language processing[7-10], computer vision[11,12], deep learning[13-15], and reinforcement learning. Reinforcement learning is often used in virtual environments like games and is crucial in advertising to evaluate the effects of initialization and exploration processes. Our system includes three key modules: offline search advertising simulation, offline reinforcement learning, and online strategy optimization. The offline simulation module simulates strategy impacts, such as ranking outcomes and user behaviors, enabling safe exploration of strategies. It produces numerous training samples for reinforcement learning, records contexts, simulates policy functions, and predicts metrics like ad views and clicks. This helps maintain user experience and platform revenue while simplifying complex real-world scenarios. In the design of the reward function, if the user clicks and platform revenue are optimized, the reward function can be designed as:

$$r(s_t, a_t) = CTR \cdot \text{click\_price} + \delta \cdot CTR \quad (3)$$

Among them, δ is the adjustment factor, which is used to adjust the balance between click-through rate and deduction.

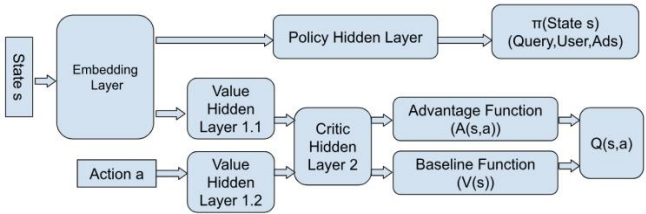

Figure 1. Strategy optimization system framework

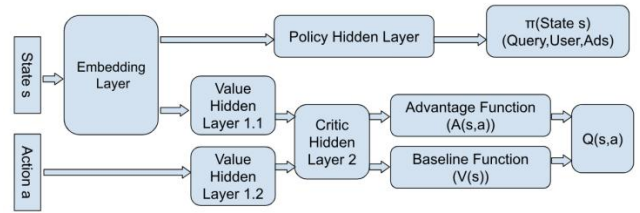

Figure 2. DDPG network structure

The offline reinforcement learning module optimizes strategies using simulated data to initialize the strategy model, employing an off-policy model with a continuous action space. We use a Deep Deterministic Policy Gradient (DDPG) model with an Actor-Critic structure and feature encoding through embedding layers. This module focuses on critical actions and supports asynchronous learning with multiple agents, enhancing strategy generation and network updates.

In strategy optimization, the system adjusts the balance between click-through rate and cost using a reward function that considers both user clicks and platform revenue. Calibration methods like Isotonic regression ensure that simulated outcomes closely match real online responses.

## V. MODEL OPTIMIZATION

Although in the process of policy optimization learning, we used an offline simulation model to explore the policy space and perform reward calibration on the estimated results, the simulation results do not represent the real behavior of users. Because the user's behavior will be affected by other environmental factors, and as mentioned in the simulation system, the simulation system still does not consider many factors, such as the advertiser's budget, bids and other information, which the simulation system cannot obtain. Serialized simulation results. Therefore, the strategy optimization algorithm needs to learn online based on real online feedback.

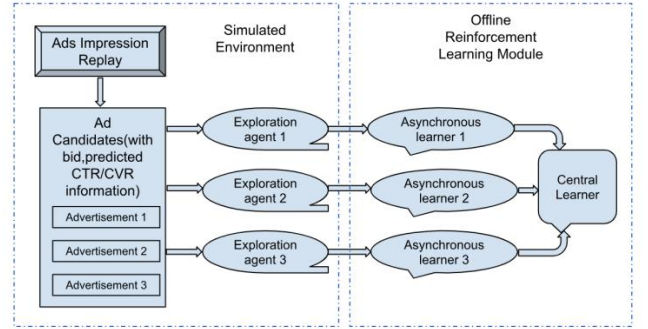

Figure 3. Offline DDPG learning process framework

For online learning methods, we use the Evolution Strategy method for online strategy updates. For a given ranking strategy model $\pi_\theta(s_t)$, Evolution Strategy performs strategy exploration and model updating by performing the following two steps: (1) Adding Gaussian noise to the model parameter space θ to generate exploration action a; (2) Statistically different The reward results obtained by the strategy under noise, and the network parameters are updated based on the results. Suppose we perturb the parameter space n times to generate the perturbed parameter space $\Theta_\pi = \theta_\pi + \epsilon_1, \theta_\pi + \epsilon_2, \ldots, \theta_\pi + \epsilon_n$. The actual reward on the corresponding line is $R_i$. Then the parameter update method is $\theta'_\pi = \theta_\pi + \eta \frac{1}{n\sigma} \sum_{i=1}^{n} \bar{R}_i \epsilon_i$, where η represents the learning rate. Using Evolution Strategy to update model parameters has three advantages. First of all, Evolution Strategy is a derivative-free update method. Using this update method can avoid the calculation amount caused by calculating gradients; secondly, under the distributed parameter-serving framework, each worker only needs to The reward value can be passed to the parameter-server, which can greatly reduce the demand for network bandwidth for online learning; finally, this method can calculate the reward as a whole in one episode without having to consider the impact of reward sparsity in the state transfer process on the algorithm. influence, thereby achieving the overall optimization effect based on the browsing sequence.

---

ALGORITHM 1: Asynchronous DDPG Learning

**Input:** Simulated transition tuple set $\Gamma$ in the form $\psi = <s_t, a_t, r_t, s_{t+1}>$

**Output:** Strategy Network $\pi_{\theta_\pi}(s_t)$

Initialize critic network $Q_{\theta_Q}(s_t, a_t)$ with parameter $\theta_Q$ and actor network $\pi_{\theta_\pi}(s_t)$ with parameter $\theta_\pi$;

Initialize target network Q', π' with weights $\theta_{Q'} \leftarrow \theta_Q$, $\theta_{\pi'} \leftarrow \theta_\pi$;

**repeat**

Update network parameters $\theta_Q, \theta_{Q'}, \theta_\pi$ and $\theta_{\pi'}$ from parameter server;

Sampling subset $\Psi = \{\psi_1, \psi_2, \ldots, \psi_m\}$ from $\Gamma$;

For each $\psi_i$, calculate $Q^* = r_t + \gamma \cdot Q'(s_{t+1}, \pi'(s_t))$;

Calculate critic loss $L = \sum_{\psi_i \in \Psi} \frac{1}{2} \cdot (Q^* - Q(s_t, a_t))^2$;

Compute gradients of Q with respect to $\theta_Q$ by $\nabla_{\theta_Q} Q = \frac{\partial L}{\partial \theta_Q}$;

Compute gradients of π with respect to $\theta_\pi$ by

$\nabla_{\theta_\pi} \pi = \sum_{\psi_i \in \Psi} \frac{\partial Q(s_t, \pi(s_t))}{\partial \pi(s_t)} \cdot \frac{\partial \pi(s_t)}{\partial \theta_\pi} = \sum_{\psi_i \in \Psi} \frac{\partial A(s_t, \pi(s_t))}{\partial \pi(s_t)} \cdot \frac{\partial \pi(s_t)}{\partial \theta_\pi}$;

Send gradients $\nabla_{\theta_Q} Q$ and $\nabla_{\theta_\pi} \pi$ to the parameter server;

Update $\theta_Q$ and $\theta_\pi$ with $\nabla_{\theta_Q} Q$ and $\nabla_{\theta_\pi} \pi$ for each global N steps by gradients method;

Update $\theta_{Q'}$ and $\theta_{\pi'}$ by $\theta_{Q'} \leftarrow \theta_{Q'} + (1-\tau)\theta_Q$, $\theta_{\pi'} \leftarrow \theta_{\pi'} + (1-\tau)\theta_\pi$;

**Until** Convergence;

## VI. EXPERIMENT ANALYSIS AND DISCUSSION

We address several key questions through experiments: Can the model converge to the optimal solution? How do different network architectures and parameter designs affect model convergence? What gains does online updating provide in enhancing the model's online performance?

To answer the first two questions, we represent the search context feature, s, simply using the query word ID. On the offline simulation platform, by employing a sliding window search over the parameter set a, we identify the optimal parameter values for the sorting function. We then compare these values with those obtained from DDPG to assess method convergence. Figures 4 and 5 showcase the training convergence under various model configurations, detailed in Table 1. The results indicate that: (1) The dueling architecture significantly boosts model convergence by distinguishing between the reward function (value function) and the advantage function (advantage function); (2) A larger training dataset size (batch size) benefits convergence due to the high variance in data.

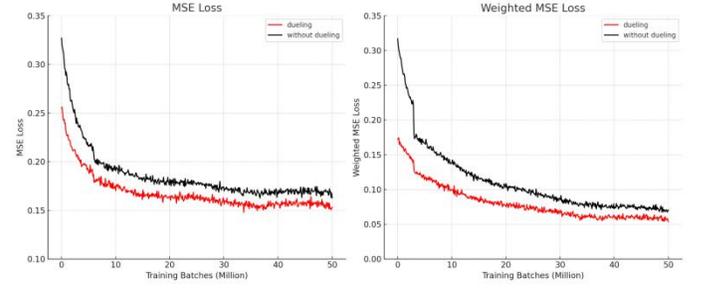

Figure 4. Effect of using dueling structure on convergence

For question three, we put the policy model learned by DDPG online for 2% traffic testing, and used ES to update the policy. The experiment lasted for 4 days, mainly comparing the changes in CTR, PPC, and RPM indicators. The experimental results are shown in the figure below.

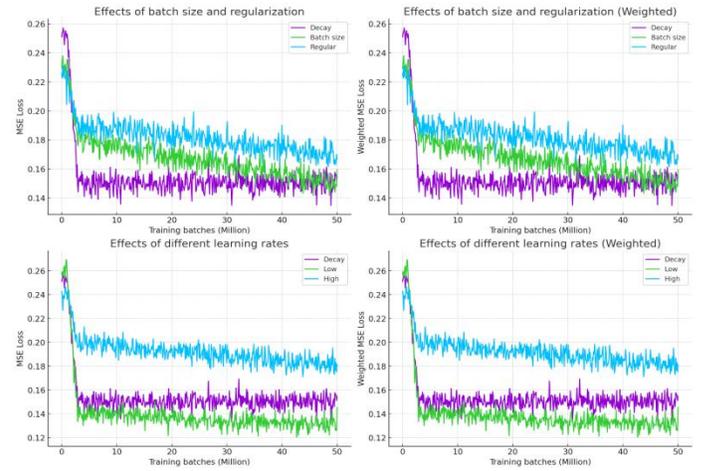

Figure 5. The impact of using different attenuation factors and different batch training data sets (batch size) on convergence

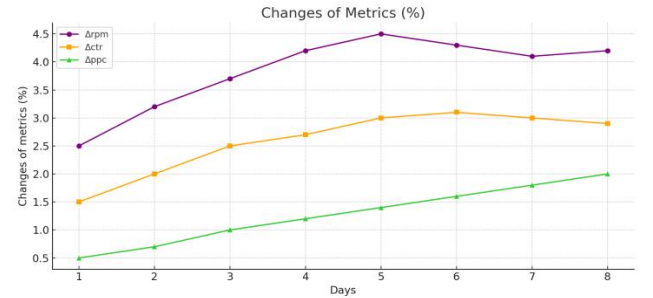

Figure 6. Online ES effect change trend

## VII. CONCLUSION

This paper emphasizes the success of integrating reinforcement learning with GSP auction models to optimize search ad placements, enhancing platform revenue, user, and advertiser satisfaction, and supporting a sustainable advertising ecosystem. Future research could expand this model to various digital advertising platforms and adapt it to changing market dynamics. The study's main limitations are its dependence on high-quality simulation data for initial training and the necessity for continuous adjustments to keep the model accurate and relevant. Investigating adaptive learning rates and other reinforcement learning algorithms may help overcome these challenges, leading to more autonomous and efficient advertising systems.

REFERENCES

[1] D. Li, Z. Zhang, B. Alizadeh, Z. Zhang, N. Duffield, M. Meyer, C. M. Thompson, H. Gao, and A. H. Behzadan, "A reinforcement learning-based routing algorithm for large street networks," International Conference on Runtime Verification, vol. 38(2), pp. 183–215, 2024.

[2] Q. Ning, W. Zheng, H. Xu, A. Zhu, T. Li, Y. Cheng et al., "Rapid segmentation and sensitive analysis of CRP with paper-based microfluidic device using machine learning," Analytical and Bioanalytical Chemistry, vol. 414(13), pp. 3959–3970, 2022.

[3] M. Nagao, C. Yao, T. Onishi, H. Chen, A. Datta-Gupta, and S. Mishra, "An efficient deep learning-based workflow for CO2 plume imaging considering model uncertainties with distributed pressure and temperature measurements," International Journal of Greenhouse Gas Control, vol. 132, p. 104066, 2024.

[4] X. Shen, Q. Zhang, H. Zheng, and W. Qi, "Harnessing XGBoost for Robust Biomarker Selection of Obsessive-Compulsive Disorder (OCD) from Adolescent Brain Cognitive Development (ABCD) data," ResearchGate 10.13140/RG.2.2.12017.70242, May 2024.

[5] X. Xie, H. Peng, A. Hasan, S. Huang, J. Zhao, H. Fang, W. Zhang, T. Geng, O. Khan, and C. Ding, "Accel-gcn: High-performance gpu accelerator design for graph convolution networks," 2023 IEEE/ACM International Conference on Computer Aided Design (ICCAD), pp. 01–09, 2023.

[6] H. Peng, S. Zhou, Y. Luo, N. Xu, S. Duan, R. Ran, J. Zhao, S. Huang, X. Xie, C. Wang, et al., "Rrnet: Towards relu-reduced neural network for two-party computation based private inference," CoRR, 2023.

[7] Y. Zhou, T. Shen, X. Geng, G. Long, and D. Jiang, "ClarET: Pre-training a Correlation-aware context-to-Event Transformer for event-centric generation and classification," Proceedings of the 60th Annual Meeting of the Association for Computational Linguistics, vol. 1, pp. 2559–2575, 2022.

[8] W. Lyu, S. Zheng, L. Pang, H. Ling, and C. Chen, "Attention-Enhancing Backdoor Attacks Against BERT-based Models," Findings of the Association for Computational Linguistics: EMNLP 2023, pp. 10672–10690, 2023.

[9] Y. Zhou, X. Geng, T. Shen, W. Zhang, and D. Jiang, "Improving zero-shot cross-lingual transfer for multilingual question answering over knowledge graph," Proceedings of the 2021 Conference of the North American Chapter of the Association for Computational Linguistics: Human Language Technologies, pp. 5822–5834, 2021.

[10] W. Lyu, S. Zheng, T. Ma, and C. Chen, "A Study of the Attention Abnormality in Trojaned BERTs," Proceedings of the 2022 Conference of the North American Chapter of the Association for Computational Linguistics: Human Language Technologies, pp. 4727–4741, 2022.

[11] Z. Li, B. Guan, Y. Wei, Y. Zhou, J. Zhang, and J. Xu, "Ground Truth Image Creation with Pix2Pix Image-to-Image Translation," CoRR, 2024.

[12] B. Dang, W. Zhao, Y. Li, D. Ma, Q. Yu, and E. Y. Zhu, "Real-Time pill identification for the visually impaired using deep learning," arXiv preprint arXiv:2405.05983, 2024.

[13] W. Xu, J. Chen, Z. Ding, and J. Wang, "Text sentiment analysis and classification based on bidirectional Gated Recurrent Units (GRUs) model," arXiv preprint arXiv:2404.17123, 2024.

[14] W. Lyu, X. Lin, S. Zheng, L. Pang, H. Ling, S. Jha, and C. Chen, "Task-Agnostic Detector for Insertion-Based Backdoor Attacks," CoRR, 2024.

[15] Z. An, X. Wang, T. T. Johnson, J. Sprinkle, and M. Ma, "Runtime monitoring of accidents in driving recordings with multi-type logic in empirical models," International Conference on Runtime Verification, pp. 376–388, 2023.